\documentclass[journal]{IEEEtran} 
\usepackage[T1]{fontenc}
\usepackage{lmodern}
\usepackage{amsmath,amsfonts}
\usepackage[linesnumbered,ruled]{algorithm2e}
\usepackage{booktabs}
\usepackage{makecell}
\usepackage{multirow}
\usepackage{tabularx} 
\usepackage{array}
\usepackage[caption=false,font=normalsize,labelfont=sf,textfont=sf]{subfig}
\usepackage{textcomp}
\usepackage{stfloats}
\usepackage{url}
\usepackage{verbatim}
\usepackage{graphicx}
\usepackage{hyperref}
\usepackage{xcolor}
\usepackage{eso-pic}
\hypersetup{
	colorlinks=true,
	linkcolor=blue,
	citecolor=blue,
	urlcolor=blue
}

\usepackage{cite}
\begin{document}

\AddToShipoutPictureBG*{
	\AtPageUpperLeft{
		\put(0,-40){ 
			\makebox[\paperwidth][c]{
				\parbox{\textwidth}{
					\centering
					\footnotesize 
					This work has been submitted to the IEEE for possible publication. \\ 
					Copyright may be transferred without notice, after which this version may no longer be accessible.
				}
			}
		}
	}
}

\title{Towards Universal Physical Adversarial Attacks via a Joint Multi-Objective and Multi-Model Optimization Framework}

\author{Ziyang Liu, Hongyuan Wang, Zijian Wang, Yinxi Lu, Yunzhao Zang, Zhiqiang Yan, and Qianhao Ning

\thanks{Manuscript created May, 2026. (Corresponding authors: Zhiqiang Yan, Qianhao Ning)}

\thanks{Ziyang Liu, Zhiqiang Yan, Qianhao Ning, Zijian Wang, Yinxi Lu, Yunzhao Zang, and Hongyuan Wang are with the Research Center for Space Optical Engineering, Harbin Institute of Technology, Harbin 150001, China. Hongyuan Wang is also with the Zhengzhou Research Institute, Harbin Institute of Technology, Zhengzhou 450000, China. (e-mail: meyondlzy@stu.hit.edu.cn, yanzhiqiang@hit.edu.cn, ningqh@hit.edu.cn).}%

}


\maketitle

\begin{abstract}

Physical adversarial attacks often overfit single surrogate models and optimization objectives. While ensemble attacks can mitigate this, existing methods struggle with severe gradient conflicts within restricted physical texture spaces, significantly degrading cross-model transferability. To bridge this gap, this paper proposes a Joint Multi-Objective and Multi-Model Optimization Framework (JMOF) that leverages quantitative similarity analysis to select the optimal surrogate model ensemble. Within JMOF, a dual-level mechanism jointly suppresses prediction outputs and flattens intermediate feature distributions, balancing attack efficiency with deep generalization. Additionally, an Orthogonal Gradient Alignment (OGA) strategy resolves cross-model gradient conflicts, transforming mutually repulsive gradients into synergistic optimization directions. Extensive simulated and real-world experiments demonstrate that JMOF outperforms state-of-the-art baselines against diverse black-box detectors. Crucially, JMOF exhibits substantial cross-vision-task generalization, generating attacks capable of simultaneously deceiving object detection and semantic segmentation or monocular depth estimation models. This research advances the generalization limits of physical adversarial attacks, providing a robust framework for evaluating visual AI vulnerabilities in real-world deployments.

\end{abstract}


\section{Introduction}
\label{sec1}

Deep Neural Networks (DNNs) have achieved unprecedented success in various computer vision tasks \cite{Ren2016faster,zou2023object,Liu2023survey}, such as object detection, but numerous studies have also demonstrated their high vulnerability to adversarial attacks \cite{2024Wei,Guo2025beyond}. Traditional digital adversarial attacks usually deceive models by adding imperceptible perturbations to images \cite{Goodfellow2015}. However, digital attacks that directly manipulate input images are often impractical in real-world scenarios. In contrast, physical adversarial attacks deployed on target surfaces pose a significantly higher practical threat. Nevertheless, transferring these digital perturbations into the complex physical world to achieve robust attacks, such as 3D adversarial camouflage, remains a formidable challenge \cite{Hull2024renderbender}.

\begin{figure}[t]
	\centering
	\includegraphics[width=3.3 in]{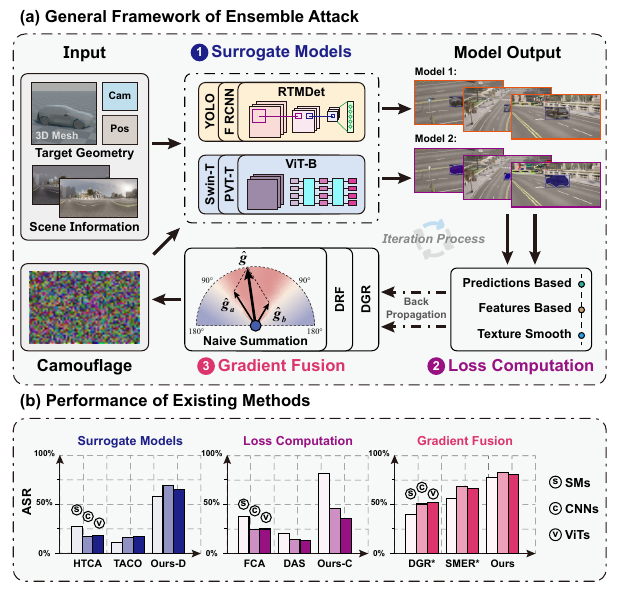}
	\caption{The general framework of existing ensemble attack methods, alongside a performance comparison of representative methods and gradient fusion strategies. Ours-D and Ours-C are ablated variants of our method; the former employs only a detection loss function, whereas the latter utilizes the same surrogate model (YOLOv3) as FCA and DAS. The asterisk (*) denotes gradient fusion methods implemented by porting gradient strategies from digital attacks into our physical attack workflow.}
	\label{fig1}
\end{figure}

Research on physical adversarial attacks has long been hindered by two critical bottlenecks. The first is the domain shift from the digital space to the physical world \cite{NPAC,Liu2025TGRS}. Specifically, complex variations in the physical environment, including lighting, viewpoints, occlusions, and distances, can easily degrade the effectiveness of adversarial perturbations. In recent years, the introduction of differentiable rendering techniques \cite{Xiao2019meshadv} and Expectation Over Transformation (EoT) \cite{EOT} has provided relatively mature solutions to this physical robustness challenge. However, a more fatal and deep-rooted problem lies in adversarial transferability. When deployed against black-box models with unknown structures and parameters, the attack performance of adversarial camouflage often declines precipitously \cite{HTCA}.

As illustrated in \hyperref[fig1]{Fig.~\ref{fig1}}-(a), generating adversarial camouflage is essentially a dynamic optimization process: learning from surrogate models, exploiting vulnerabilities via loss functions, and updating perturbations via gradient fusion. Based on this logic, we observe that existing physical attack methods exhibit severe limitations across these three core stages when targeting black-box transferability:

1) Overfitting to a single surrogate model: Relying on a single surrogate model (e.g., the YOLO series) causes the generated camouflage to overfit specific decision boundaries, leading to rapid failure against real-world heterogeneous networks. While ensemble attacks can mitigate this by learning richer cross-model representations \cite{Pi2023improving}, achieving joint optimization across diverse architectures remains highly challenging due to the restricted parameter space of physical attacks.

2) Lack of synergy between shallow result suppression and deep feature disruption: Existing methods typically employ a singular attack mode. Shallow attacks \cite{FCA} efficiently suppress final outputs but overfit specific decision boundaries, whereas deep attacks \cite{DAS} perturb shared semantic features for better transferability but often fall into optimization pitfalls due to the lack of rigid output supervision. Without effectively synergizing these two strategies, current attacks struggle to achieve both high efficiency and deep generalizability.

3) Gradient conflicts in multi-source fusion: When integrating multiple models and objectives, existing strategies mostly adopt equal-weight summation \cite{TACO} or compromise-based fusion \cite{DRF,DGR}. However, the limited texture parameters in physical attacks amplify severe conflicts and oscillations in heterogeneous gradients during optimization. Both directional cancellation and energy loss in these naive fusion methods inevitably mask the adversarial demands of robust models, leading to attack degradation or unidirectional overfitting.

In summary, constrained by the methodological design relying on a single surrogate model and a single attack objective, as well as the restricted parameter space inherent in physical adversarial attacks, existing methods are effective only under specific task settings, rapidly losing their deceptive capabilities when deviating from these presets. Not to mention that when an attack attempts to cross the boundaries of visual tasks, such as transferring from detection to segmentation or depth estimation tasks, the semantic gap in gradients caused by differences in underlying optimization objectives and network architectures expands drastically, resulting in a substantial performance degradation for existing joint optimization strategies. This landscape clearly demonstrates that a massive gap still exists between the capability boundaries of current physical adversarial attacks and the real-world deployment demands involving black-box and multi-task scenarios, making the stride toward universal attacks highly challenging.

To address the aforementioned limitations, this paper proposes a Joint Multi-Objective and Multi-Model Optimization Framework (JMOF). This framework is capable of generating universal physical adversarial attacks that exhibit high physical robustness and transferability across different models and tasks. The primary contributions of this work are summarized as follows:

1) We analyze the failure modes of joint multi-model adversarial optimization, and reveal a critical trade-off in gradient similarity of surrogate model: overly aligned gradients limit representation diversity, while highly divergent gradients induce severe conflicts. Based on this, we define an effective similarity regime that guides surrogate model selection.

2) We propose a dual-level multi-objective attack mechanism, which jointly suppresses prediction outputs and disrupts intermediate feature representations via targeted output suppression and feature distribution flattening, improving both attack strength and transferability.

3) Within JMOF, we introduce an Orthogonal Gradient Alignment (OGA) module to resolve gradient conflicts and enforce coherent optimization, enabling adversarial perturbations to converge toward shared vulnerable regions across heterogeneous models.

4) Extensive digital-to-physical experiments demonstrate that our method achieves superior transferability, outperforming prior SOTA methods across black-box detectors, and further exhibits strong cross-task generalization under joint optimization across diverse vision tasks such as semantic segmentation and monocular depth estimation, maintaining high detection attack performance while simultaneously disrupting these tasks.

\section{Related Works}
\label{sec2}

\subsection{Physical Adversarial Attacks}
\label{sec21}

The inception of physical adversarial attacks can be traced back to the work of Kurakin et al. \cite{I_FGSM}, who first demonstrated the feasibility of printing digital adversarial examples into the real world. Subsequently, to enhance deployment flexibility, researchers proposed adversarial patch techniques. For instance, Thys et al. \cite{Thys2019fooling} successfully implemented physical invisibility against YOLO object detectors using printed stickers. However, this strategy of directly printing 2D planar images with adversarial perturbations as patches harbors inherent physical flaws. When confronting complex viewpoint transformations and object deformations in the real world, 2D patches are highly susceptible to failure in multi-view detection. To overcome this limitation, research focus has gradually shifted toward 3D physical adversarial attacks targeting 3D entities in the physical world, such as the 3D surface texture of vehicles. For example, Xiao et al. \cite{Xiao2019meshadv} first proposed using differentiable rendering techniques to backpropagate the loss gradients of object detectors directly to the texture attributes of a 3D mesh. Furthermore, the method proposed by Huang et al. \cite{UPC} optimizes the 3D texture of the target directly within the 3D rendering pipeline, thereby enhancing the attack effectiveness across all viewpoints.

While 3D adversarial camouflage poses a significant threat, it faces two primary bottlenecks: physical robustness and cross-model transferability \cite{Cheng2024feature,Li2024survey}. Recent advances in differentiable rendering, such as the Optical Imaging Physical Process
Drven Differentiable Renderer (OPDR) \cite{NPAC}, have substantially bridged the digital-to-physical domain gap, largely mitigating the robustness issue. Consequently, cross-model transferability has emerged as the most critical hurdle restricting the practical deployment of physical attacks.

\subsection{Transferable Attacks}
\label{sec22}

In recent years, to enhance the cross-model transferability of adversarial attacks, deep feature-based disruption strategies, such as FIA \cite{FIA}, have been proposed. These strategies indicate that attack methods that rely on shallow outputs of model predictions are prone to overfitting the decision boundaries of specific surrogate models, leading to frequent failures when confronting heterogeneous networks \cite{ILA, TGR}. Nevertheless, despite varying network architectures, different models often extract highly similar intermediate semantic features or attention maps when processing the same visual task \cite{Kornblith2019similarity,Raghu2021vision}. Therefore, accurately perturbing these critical high-level features can significantly improve the cross-model transferability of adversarial perturbations. For instance, Zhang et al. \cite{EPA} introduced the EPA method, which uses a Jacobian matrix to map physical perturbation pixels to key target features, aligning patch-level optimization gradients with the disruption direction of deep key features and thereby directing the attention of heterogeneous models toward the target. Additionally, the hierarchical feature disruption module proposed in the HTCA method by Wang et al. \cite{HTCA} extracts and disrupts shared spatial structural features of targets across multiple network levels and scales, thereby enhancing the black-box generalization capability of adversarial camouflage in the physical world.

As illustrated in \hyperref[fig1]{Fig.~\ref{fig1}}-(b), shallow attacks (e.g., FCA) efficiently suppress specific model outputs but struggle with heterogeneous generalization. Conversely, deep attacks (e.g., DAS) perturb shared high-level semantic features, inherently possessing stronger cross-architecture transferability but lacking explicit objective guidance at the output level. Recognizing their functional complementarity, this paper proposes a synergistic integration of shallow result suppression and deep feature disruption. This dual-level strategy overcomes the trade-offs of single-level attacks, providing a novel paradigm for generating universal physical camouflage.

\begin{figure*}[t]
	\centering
	\includegraphics[width=1.0\textwidth]{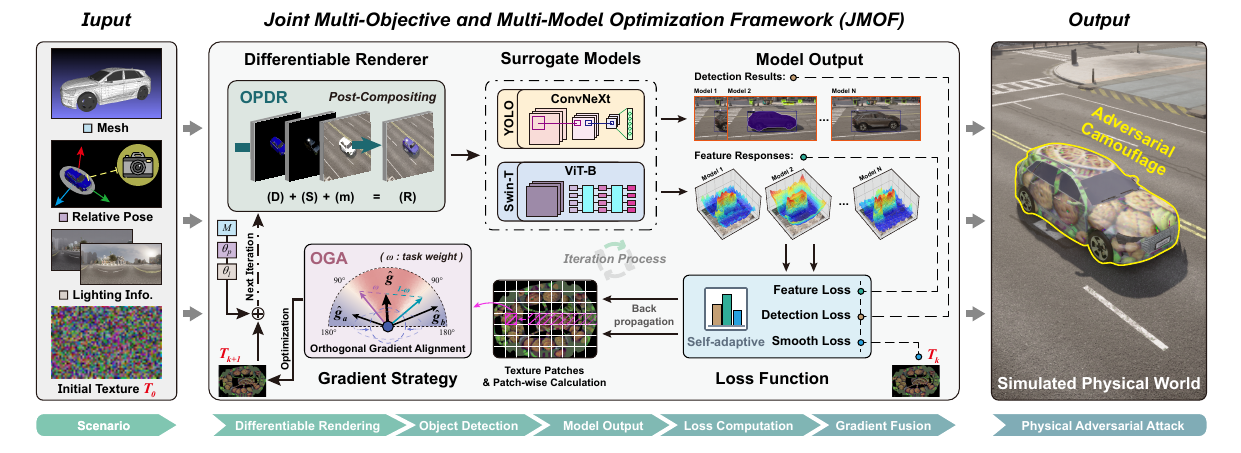}
	\caption{The workflow of JMOF, illustrating the optimization progression from $T_k$ to $T_{k+1}$ within the framework.}
	\label{fig2}
\end{figure*}

\subsection{Ensemble Attacks}
\label{sec23}

Despite alleviating overfitting, previous strategies remain constrained by the single-surrogate model paradigm, whose limited parameter space cannot sufficiently capture the universal cognitive logic of the underlying visual task \cite{Andrew2019, Kaleel2021}. To break this information barrier, ensemble attack architectures are essential for guiding perturbations toward shared vulnerability spaces across diverse models. Dimitriu et al. \cite{TACO} developed a simple joint multi-model method, TACO, based on the YOLO series of networks. However, their selection of surrogate models relied on intuition, lacking quantitative profiling of gradient similarity and feature overlap among models, resulting in poor attack transferability. As shown in \hyperref[fig1]{Fig.~\ref{fig1}}-(b), when we replace the surrogate models of TACO with significantly different heterogeneous models, its transferability is inferior to that of HTCA, which is based on single-model optimization. This demonstrates that gradient conflicts and directional cancellations arising from insufficient guidance can severely limit the transferability of ensemble attacks, highlighting the critical importance of a gradient fusion mechanism capable of resolving such conflicts \cite{Xue2024}.

In the unconstrained digital domain, ensemble methods like DRF \cite{DRF} and DGR \cite{DGR} address gradient conflicts by passively discarding conflicting components. However, this compromise strategy dissipates optimization energy and irreversibly loses crucial feature preferences, leading to gradient decay. Even recent advanced methods, such as SMER \cite{SMER} and NAMEA \cite{NAMEA}, often suffer from performance degradation in 3D physical scenarios, As shown in \hyperref[fig1]{Fig.~\ref{fig1}}-(b), due to intense optimization oscillations and semantic fragmentation.

Currently, within the highly constrained parameter space of the physical domain, there is no joint optimization framework capable of accurately quantifying and actively resolving gradient conflicts across complex heterogeneous networks, or even across multiple visual tasks. Therefore, our proposed OGA strategy abandons passive discarding mechanisms, endowing joint training with the capability for dynamic perception and active restructuring. By rectifying the repulsion directions among heterogeneous gradients, it converts conflicting optimization potential energy into synergistic attack kinetic energy, thereby guiding physical adversarial camouflage to accurately converge upon the shared vulnerability space across architectures and tasks.

\section{Methodology}
\label{sec3}

In this section, we introduce our proposed method by focusing on physical adversarial attacks against object detection.

\subsection{Problem Definition}
\label{sec31}

\begin{figure*}[t]
	\centering
	\includegraphics[width=0.9 \textwidth]{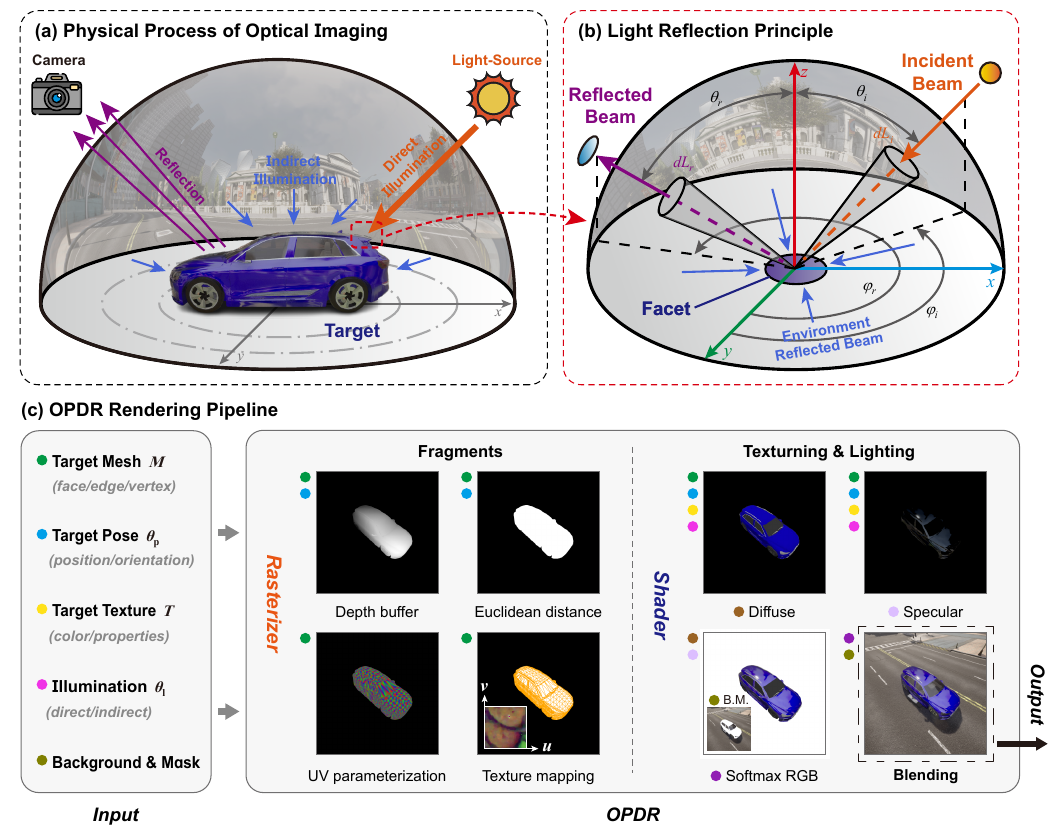}
	\caption{Illustration of the physical process of optical imaging and the rendering pipeline of OPDR.}
	\label{fig3}
\end{figure*}

The workflow of the proposed JMOF method is illustrated in \hyperref[fig2]{Fig.~\ref{fig2}}. For physical adversarial attacks on object detection, the dataset is collected using the CARLA simulator \cite{CARLA}. At each optimization step $k$, an input image $I^k$ containing the background and the target to be camouflaged is provided. To accurately reconstruct the physical reflection information of the camouflaged target during the optimization process utilizing OPDR, the relative pose $\theta_p^k$ between the target and the camera, along with the global illumination information $\theta_l^k$ for each image, are simultaneously recorded from the CARLA simulator during dataset generation. Throughout the optimization process, the target to be camouflaged remains unique, with all images sharing a common 3D mesh $M$ and an initial surface texture $T_0$. The optimization objective of this task is to enable the camouflaged target to simultaneously deceive multiple jointly trained object detection models.

Therefore, in the $k$-th optimization step of an epoch, the input information for the algorithm is defined as $\{I^k, \theta_p^k, \theta_l^k, M, T^k\}$. Based on this information, as shown in \hyperref[eq1]{Eq.~(\ref{eq1})}, the renderer $R$ can accurately reconstruct the rendering result of the physical entity when the adversarial texture is applied to the given image. Subsequently, this rendered result is superimposed back onto the original image utilizing a mask $m$ to obtain the adversarial example $I_{adv}^k$ for the physical attack.

\begin{equation}
	I_{adv}^k=m\odot\mathbf{R}\left(\theta_p,\theta_l,M,T^k\right)+\left(1-m\right)\odot I^k
	\label{eq1}
\end{equation}

\noindent The adversarial example $I_{adv}^k$ is then simultaneously fed into a surrogate model ensemble $D=\{D_1,D_2,\dots,D_N\}$ comprising $N$ object detection networks with distinct architectures, yielding the intermediate feature responses $F^k$ and the detection outputs $O^k$ of the models. This information, alongside the current texture $T^k$, is subsequently incorporated into the loss function set $L=\{L_1,L_2,\dots,L_N\}$ for computation. Given that optimization results are highly sensitive to gradient conflicts and internal energy dissipation within the highly restricted parameter space of physical adversarial attacks, we introduce the OGA strategy during backpropagation. Initially, the independent backpropagation gradients $g_{i}^{k}=\nabla_{T_{k}}\mathcal{L}_{i}$ of each surrogate model with respect to the current texture parameter $T^k$ are computed. Following this, OGA dynamically quantifies the repulsion angle between heterogeneous gradients and proactively performs orthogonal projection and directional realignment on the conflicting gradient components. This mechanism synthesizes a synergistic global optimization gradient $g^k$ without discarding the feature preferences of individual models. The schematic formulation of this process is represented in \hyperref[eq2]{Eq.~(\ref{eq2})}.
\begin{equation}
	g^k=\mathrm{OGA}(g_1^k,g_2^k,...,g_N^k)
	\label{eq2}
\end{equation}

\noindent Finally, the optimizer iteratively updates the current texture using this synergistic global gradient $g^k$, obtaining the input texture $T^{k+1}$ for the subsequent optimization step. In summary, the algorithm and process of the proposed adversarial attack against object detection models can be simplified as \hyperref[eq3]{Eq.~(\ref{eq3})}.

\begin{equation}
	T^{k+1}=\arg\min_{T^k}\left(\mathrm{OGA}\left(\nabla_{T^k}\mathcal{L}\left(O^k,T^k,G^k\right)\right)\right)
	\label{eq3}
\end{equation}

\subsection{Differentiable Renderer}
\label{sec32}

The principle of OPDR is based on the physical process of optical imaging, As illustrated in \hyperref[fig3]{Fig.~\ref{fig3}}-(a), (b) and \hyperref[eq4]{Eq.~(\ref{eq4})}, simulating both direct light source illumination and indirect environmental reflections using a Bidirectional Reflectance Distribution Function (BRDF) \cite{Yan2022physics,Yan2022imaging}. 

\begin{equation}
	f_\mathrm{r}(\theta_i,\varphi_i,\theta_r,\varphi_r,\lambda)=\frac{\mathrm{d}L_r\left(\theta_i,\varphi_i,\theta_r,\varphi_r,\lambda\right)}{\mathrm{d}L_i\left(\theta_i,\varphi_i,\lambda\right)}
	\label{eq4}
\end{equation}

\noindent In this equation, $\theta_i$ represents the incident zenith angle, $\phi_i$ denotes the incident azimuth angle, $\theta_r$ stands for the reflected zenith angle, $\phi_r$ specifies the reflected azimuth angle, $\lambda$ is the incident light wavelength, $dL_r$ indicates the reflected spectral radiance of the microfacet in the $(\theta_r, \phi_r)$ direction, and $dL_i$ signifies the incident spectral irradiance on the microfacet.

As illustrated in \hyperref[fig3]{Fig.~\ref{fig3}}-(c), the OPDR pipeline differentiably maps the 3D mesh $M$, relative pose $\theta_p$, global illumination $\theta_l$, and surface texture $T$ into a 2D rendered image. Since the comprehensive mathematical formulations of the BRDF integrations and the detailed rendering pipeline have been thoroughly elaborated in our previous work \cite{NPAC}, we omit the redundant derivations here and focus on its core differentiable mechanisms.

In traditional non-differentiable rendering, discrete rasterization introduces gradient discontinuities at screen-space boundaries, depth occlusions, and texture seams, severely disrupting the gradient flow required for adversarial optimization \cite{pytorch3d}. To address these issues, OPDR adopts a probabilistic blending strategy based on soft rasterization. It converts discrete stepping at boundaries into fuzzy, probabilistic transitions via a differentiable Sigmoid function, replaces traditional hard occlusion testing with probabilistic transparency blending, and uses bilinear interpolation for texture sampling to preserve sub-pixel gradient continuity. These mechanisms ensure stable gradient propagation throughout the entire renderer, thereby robustly supporting the backward optimization of physical adversarial camouflage.

\subsection{Surrogate Models}
\label{sec33}

Existing ensemble attacks often select surrogate models intuitively or randomly, leading to homogenized feature extraction logic and convergence into local minima. To prevent this severe bias and elevate cross-model transferability, it is essential to construct a highly diverse surrogate model ensemble encompassing a broad spectrum of visual cognitive logic.

To overcome the limitations inherent in current rudimentary surrogate model selection strategies, we propose utilizing underlying optimization dynamics characteristics, specifically Gradient Cosine Similarity \cite{Liu2016delving,Kornblith2019similarity}, to quantify the degree of heterogeneity among different surrogate models. For a given input image and similar optimization objectives, the gradient vectors $g_a$ and $g_b$ generated by different models $D_a$ and $D_b$ through backpropagation directly reflect their spatial preferences for vulnerable target features \cite{Ren2025improving}. We define the gradient similarity $S(D_a, D_b)$ between two models as formulated in \hyperref[eq5]{Eq.~(\ref{eq5})}.

\begin{equation}
	S(D_a,D_b)=\frac{\langle g_a,g_b\rangle}{\|g_a\|_2\|g_b\|_2}
	\label{eq5}
\end{equation}

Here, $\langle \cdot \rangle$ denotes the vector inner product. A value of $S(D_a, D_b)$ closer to 1 indicates a higher consistency in the gradient update directions of the two models, signifying their reliance on highly overlapping visual features. Conversely, a similarity approaching -1 indicates substantial divergences in the cognitive logic and attention distributions of the two models during target detection, signifying profound heterogeneity.

\begin{figure}[t]
	\centering
	\includegraphics[width=3.3 in]{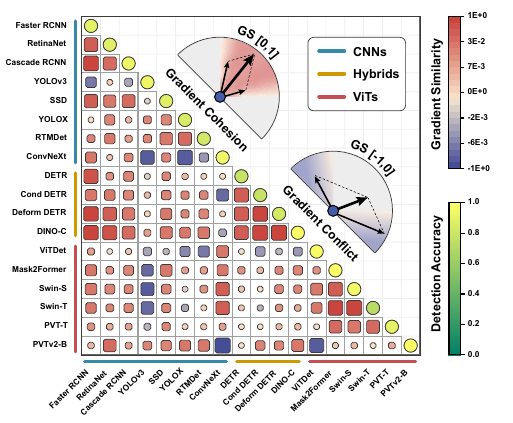}
	\caption{Gradient cosine similarity results among various models in the candidate pool. For enhanced visualization, the heatmap intensity corresponds to the extremity of the similarity values. The filled circles along the diagonal indicate the detection accuracy of each corresponding model on uncamouflaged targets.}
	\label{fig4}
\end{figure}

Based on the aforementioned model similarity metric, we employ a Greedy Selection algorithm to extract the $N$ most diverse models from a model pool $D_{pool}$ containing numerous candidate detectors, thereby constructing the joint training surrogate ensemble $D$. Initially, we calculate the average gradient similarity matrix between pairwise models within the candidate pool on a sampled data subset, as illustrated in \hyperref[fig4]{Fig.~\ref{fig4}}, and select the pair of models exhibiting the lowest similarity to form the initial ensemble. Subsequently, we evaluate the similarity of each remaining candidate model against all models within the currently selected ensemble. The model exhibiting the minimum maximum cosine similarity to the selected ensemble—representing the most incompatible network with the existing group and possessing the most unique features—is then incorporated into the ensemble. This step is repeated until the size of the surrogate model ensemble reaches the designated parameter $N$.

\subsection{Loss function}
\label{sec34}

\begin{figure}[t]
	\centering
	\includegraphics[width=3.3 in]{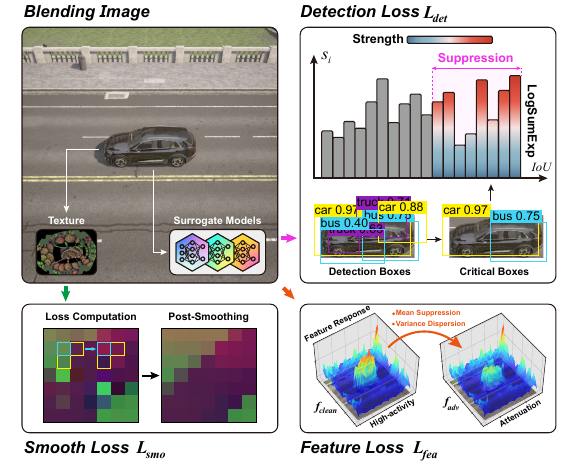}
	\caption{Visualization of the loss function design for attacking object detection models.}
	\label{fig5}
\end{figure}

To generate adversarial attacks possessing both high physical robustness and strong cross-model transferability, as illustrated in \hyperref[fig5]{Fig.~\ref{fig5}}, we design a dual-level objective comprising a Detection Loss and a Feature Loss. This leverages the explicit guidance of shallow outputs and the generalizability of deep shared semantics. Furthermore, to address real-world deployment constraints, we introduce a Smooth Loss to ensure the physical robustness of the adversarial perturbations.

\subsubsection{Detection Loss}
\label{sec341}

The detection loss $\mathcal{L}_{det}$ directly suppresses the prediction confidence of the object detector toward the camouflaged target from the output end. Deviating from traditional strategies that solely suppress the scores of specific categories, this paper adopts an indiscriminate suppression strategy. As shown in \hyperref[eq6]{Eq.~(\ref{eq6})}, among the set of bounding boxes predicted by the network, we filter out all predicted boxes whose Intersection over Union (IoU) with the Ground Truth bounding box exceeds a predefined threshold. Regardless of the object category assigned by the model, these are classified as hazardous boxes and are subsequently suppressed. To prevent gradient dispersion caused by multiple hazardous boxes, we introduce the LogSumExp function to aggregate and penalize the confidence scores $S_i$ of these hazardous boxes:

\begin{equation}
	\mathcal{L}_{\mathrm{det}}=\left(\log\sum\exp(S_i)\right)^2
	\label{eq6}
\end{equation}

\noindent This mechanism can adaptively concentrate optimization computational power on the most threatening predicted boxes.

\subsubsection{Feature Loss}
\label{sec342}

Relying exclusively on label loss at the output end is prone to causing adversarial perturbations to overfit the decision boundaries of specific models. To enhance black-box transferability, we design a feature loss $\mathcal{L}_{fea}$ to dismantle the spatial structural features shared by heterogeneous networks at the feature map level. We downsample the target mask $m$ and map it onto the multi-scale intermediate layer features $f$ extracted by the detector. For the $l$-th feature layer, the mean and variance of its features are presented in \hyperref[eq7]{Eq.~(\ref{eq7})}.

\begin{equation}
	\mu=\frac{\Sigma(f\odot m)}{\sum m},\quad\sigma^2=\frac{\sum((f-\mu)^2\odot m)}{\sum m}
	\label{eq7}
\end{equation}

Based on this, as formulated in \hyperref[eq8]{Eq.~(\ref{eq8})}, $\mathcal{L}_{fea}$ encompasses two synergistic dimensions. The first is Mean Suppression, which attenuates the overall activation values within the target region. The second is Variance Flattening, which smooths out the prominent semantic peaks within the target region.

\begin{equation}
	\mathcal{L}_{fea}=\frac{1}{L}\sum_{l=1}^{L}\log\left(1+\mu_{l}^{2}+\sqrt{\sigma_{l}^{2}+\epsilon}\right)
	\label{eq8}
\end{equation}

\noindent In this equation, a logarithmic function is employed to prevent numerical explosion and gradient runaway during the early stages of training.

\subsubsection{Smooth Loss}
\label{sec343}

In digital adversarial attacks, due to the singular attack viewpoint, adversarial perturbations frequently manifest as disordered high-frequency noise. However, when this high-frequency noise is deployed into the complex physical world, the capture of camouflaged targets by real cameras is accompanied by continuous viewpoint variations, distance scaling, and illumination interference. During this physical sampling process, these high-frequency, fragmented adversarial textures are highly susceptible to aliasing and information loss, leading to complete attack failure. Consequently, physical adversarial attacks must abandon pixel-level discrete noise and instead rely on smooth and continuous global semantic perturbations to maintain physical robustness across multiple viewpoints. To this end, as shown in \hyperref[eq9]{Eq.~(\ref{eq9})}, we introduce a Total Variation-based \cite{Rudin1992nonlinear, ACTIVE} smooth loss ($\mathcal{L}_{smo}$) during the texture parameter update to ensure the continuity of the adversarial texture in the physical space:

\begin{equation}
	\mathcal{L}_{smo}=\sum_{i,j}\sqrt{\left(T_{i+1,j}-T_{i,j}\right)^2+\left(T_{i,j+1}-T_{i,j}\right)^2}
	\label{eq9}
\end{equation}

\noindent Here, $T$ denotes the 2D adversarial texture to be optimized. $\mathcal{L}_{smo}$ is utilized to penalize drastic color jumps between adjacent pixels, eliminating high-frequency noise that easily emerges during the adversarial optimization process, thereby ensuring that the generated adversarial attack is smoother and possesses higher physical robustness.

\subsection{Gradient Fusion}
\label{sec35}

\begin{figure}[t]
	\centering
	\includegraphics[width=3.3 in]{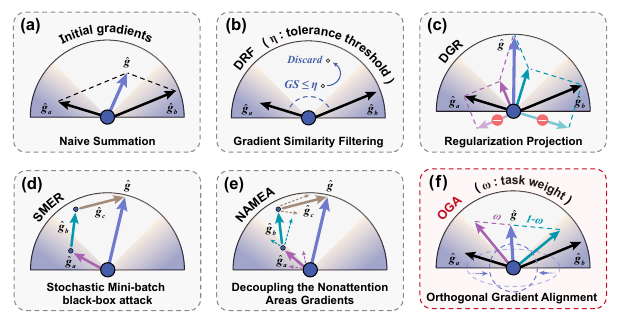}
	\caption{Schematic illustration of existing gradient fusion strategies and the OGA strategy in the presence of gradient conflicts during ensemble attacks.}
	\label{fig6}
\end{figure}

Integrating heterogeneous surrogate models inevitably leads to severe gradient conflicts and magnitude discrepancies, which are drastically amplified in the restricted parameter space of physical textures. To resolve this, we propose the OGA method. As illustrated in \hyperref[fig6]{Fig.~\ref{fig6}}, unlike passive discarding strategies, OGA utilizes orthogonal projection to extract common gradient components across architectures and actively rectifies repulsion directions. Consequently, it restructures conflicting optimization energy into highly synergistic attack kinetic energy without discarding any model's critical feature preferences.

Specifically, considering that physical adversarial attacks are deployed on complex 3D surfaces, gradient conflicts among heterogeneous models often exhibit strong spatial locality. For instance, gradient directions from different networks might be diametrically opposed in the car door region while highly consistent in the roof region. Employing a global fusion strategy akin to digital attacks tends to obliterate these local conflict features during the averaging process. Therefore, in the OGA calculation, the continuous global surface texture $T$ is first divided into $P$ non-overlapping local texture patches $T_p$. Within each local region, for the $N$ heterogeneous surrogate models, OGA extracts their respective independent local backpropagation gradients and constructs a local gradient matrix $g_p=\{g_{p,1}, g_{p,2}, \dots, g_{p,K}\}$. Subsequently, the Gram matrix $M_p = G_p^T G_p$ is computed for each local patch to accurately capture the geometric angles and magnitude correlations among cross-model gradients within that region. Following this, as shown in \hyperref[eq10]{Eq.~(\ref{eq10})}, eigenvalue decomposition is performed on $M_p$:

\begin{equation}
	\mathrm{M}_{_p}=\mathrm{V}_{_p}\Sigma_{_p}\mathrm{V}_{_p}^{T}
	\label{eq10}
\end{equation}

\noindent Here, $V_p$ is a matrix containing orthogonal eigenvectors, representing the orthogonal basis of heterogeneous gradients in space, indicating independent, non-conflicting directions. $\Sigma_p = \text{diag}(\lambda_{p,1}, \lambda_{p,2}, \dots, \lambda_{p,K})$ is a diagonal matrix, where the eigenvalues on its diagonal represent the energy magnitude of the gradients on each orthogonal component. To eliminate gradient dominance among heterogeneous models and achieve geometric alignment, OGA constructs an orthogonal transformation and energy reallocation matrix $B_p$ as formulated in \hyperref[eq11]{Eq.~(\ref{eq11})}.

\begin{equation}
	B_p=\sqrt{\bar{\lambda}_p}V_p\Sigma_p^{-1/2}V_p^T
	\label{eq11}
\end{equation}

\noindent In this equation, the gradient average is established as the scaling baseline for global gradients in OGA. This is because, in black-box adversarial scenarios, the magnitude span of heterogeneous model gradients is immense, and this average objectively reflects the current energy expectation of the entire system. This design not only forcibly scales all gradient components to a unified energy baseline, achieving orthogonal gradient alignment of heterogeneous models, but more importantly, it maintains stable overall attack kinetic energy during the later stages of optimization, specifically when the loss of a certain surrogate model plateaus and its local gradient approaches zero. This mechanism effectively prevents the attack optimization from stagnating.

After computing the patch-wise alignment matrix, these conflict-resolved gradients need to be fused into a unified update direction. However, due to the varying complexities of the underlying network architectures, the robustness of the models also differs. Direct equal-weight summation would inevitably mask the adversarial demands of complex models. Consequently, OGA introduces a dynamic task weighting mechanism $\omega$ to perform asymmetric combination based on the aligned orthogonal gradients, as expressed in \hyperref[eq12]{Eq.~(\ref{eq12})}:

\begin{equation}
	g_{p}=\sum_{k=1}^{N}\omega_{k}\cdot(B_{p}\cdot g_{p,k})
	\label{eq12}
\end{equation}

The weight vector $\omega=\{\omega_1, \omega_2, \dots, \omega_K\}$ is dynamically calculated in each iteration according to the real-time adversarial status of each model, satisfying $\sum \omega = 1$. To compensate for the inherent capability disparities among models, $\omega_K$ is set to be positively correlated with the current adversarial difficulty of the surrogate model $D_k$ and the relative proportion of its detection loss. For models that are difficult to compromise and consistently exhibit high loss values, such as Vision Transformers (ViTs) rich in high-level semantics, the algorithm adaptively assigns them larger weights. This dynamic weighting mechanism guides OGA to allocate adversarial energy more precisely toward complex models, compelling the adversarial camouflage to evolve toward vulnerable spaces posing higher-dimensional visual cognitive challenges.

\begin{figure}[t]
	\centering
	\includegraphics[width=3.3 in]{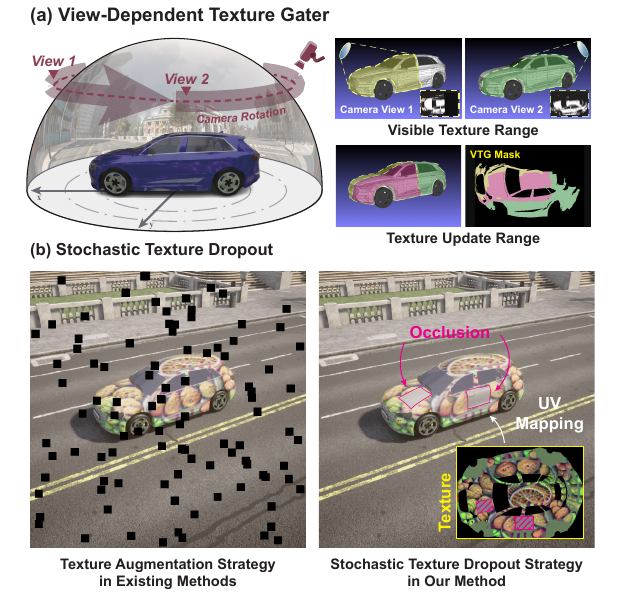}
	\caption{Principles and implementation schemes of the VTG and STD enhancement strategies.}
	\label{fig7}
\end{figure}

Furthermore, it is noteworthy that OGA not only resolves directional conflicts but also intrinsically possesses a crucial adaptive step-size decay characteristic. When severe gradient conflicts occur between two surrogate models, it indicates that the current parameters reside in an extremely steep ravine or saddle region of the loss surface \cite{Yu2020gradient}. In such critical regions, traditional rigid fusion is highly prone to causing the optimizer to skip the optimal solution in a single step due to an excessively large gradient. Following OGA's orthogonal decomposition and restructuring, strongly repulsive gradient components are compressed, thereby naturally shortening the length of the synthesized gradient in areas of intense conflict. This geometric adaptability ensures that, during joint training, when navigating these highly difficult-to-optimize conflict nodes, the algorithm explores the global optimal solution shared by multiple models at a steadier, more cautious pace, thereby fundamentally guaranteeing the stability of the physical joint adversarial attack.

\subsection{Enhancement Strategies}
\label{sec36}

\subsubsection{View-Dependent Texture Gater}
\label{sec361}

Due to the self-occlusion property of 3D entities, the detection and feature losses backpropagate gradients only to the visible local textures at the current viewpoint, whereas the physical smooth loss continuously operates on the entire global texture map. Failing to impose strict visibility constraints results in severe optimization degradation: high-intensity adversarial textures optimized in previous steps will instantly lose their adversarial gradients once occluded by viewpoint shifts. Consequently, they are erroneously smoothed away by the physical smooth loss, severely decelerating convergence.

To resolve this optimization pitfall, we introduce the View-Dependent Texture Gater (VTG), as illustrated in \hyperref[fig7]{Fig.~\ref{fig7}}-(a). During the rasterization and depth calculation phases of the OPDR pipeline, the system extracts the visibility mask of the target surface. When executing the final parameter update, VTG strictly enforces visibility constraints by severing the backpropagation links for all invisible pixels. This protects invisible adversarial textures from inadvertent degradation by the smooth loss and ensures that all valuable adversarial energy is concentrated on the most visually threatening regions under the current viewpoint, facilitating stable convergence under stringent physical-space constraints.

\subsubsection{Stochastic Texture Dropout}
\label{sec362}

During joint optimization, the high sensitivity of object detectors to specific local features often traps adversarial optimization in an energy imbalance, concentrating attack energy on a few vulnerable pixels \cite{Xie2019improving}. If these pixels are occluded in the real world, the attack collapses. While some methods attempt to alleviate this by directly adopting digital-domain 2D random erasing on rendered images, this approach is fundamentally flawed. Due to complex topological deformations and perspective foreshortening during 3D-to-2D projection, planar masking causes severe fragmentation and discontinuity when mapped onto the physical surface. This deviates from real-world occlusion patterns and introduces unreasonable gradient noise, disrupting convergence.

To break this pathological reliance and align with physical deployment laws, we propose Stochastic Texture Dropout (STD) tailored for 3D surfaces, as depicted in \hyperref[fig7]{Fig.~\ref{fig7}}-(b). Prior to forward rendering, STD randomly generates a spatial binary dropout mask with probability $p$, temporarily obscuring random blocks of the current adversarial texture $T$. This physically simulates potential real-world local occlusions. Dynamically, it forces the joint training algorithm to eschew reliance on a single local shortcut, compelling it to uncover alternative visual vulnerabilities under incomplete information. This randomized regularization uniformly distributes adversarial energy across the 3D entity, significantly strengthening physical robustness and compelling the model to learn features with greater global generalizability, thereby further enhancing cross-architecture transferability.

\begin{algorithm}[t]
	\caption{Physical Adversarial Camouflage Generation}
	\label{Algorithm1}
	\KwIn{mesh $\mathbf{M}$, input clean image set $\mathbf{I}$, relative pose $\boldsymbol{\Phi}_p$, illumination information $\boldsymbol{\Phi}_l$, mask $\mathbf{m}$, OPDR module $\mathbf{R}$, detectors $\mathbf{D}$} 
	\KwOut{Adversarial texture $T$}
	
	\tcp{Surrogate Models Selection via Cosine Similarity}
	Compute gradient cosine similarity matrix $\mathbf{S}$ for all candidate models in $\mathbf{D}$\;
	Select optimal surrogate ensemble $\mathbf{D}_{SM} \subset \mathbf{D}$ minimizing pairwise similarity in $\mathbf{S}$\;
	
	Initialize $T$ with random noise\;
	\For{the number of epochs}{
		
		Sample minibatch $I \in \mathbf{I}, \theta_p \in \boldsymbol{\Phi}_p, \theta_l \in \boldsymbol{\Phi}_l$\, $m \in \mathbf{m}$;
		
		\For{the number of steps in epoch}{
			
			$T' \leftarrow \text{STD}(T)$\;
			
			$I_{adv}, m_{VTG} \leftarrow m\odot\mathbf{R}\left(\theta_p,\theta_l,M,T'\right)+\left(1-m\right)\odot I$\;
			
			$\mathbf{O}, \mathbf{F} \leftarrow \mathbf{D}_{SM}(I_{adv})$\;
			Calculate loss $\mathcal{L}$ by \hyperref[eq6]{Eq.~(\ref{eq6})}, \hyperref[eq8]{Eq.~(\ref{eq8})} and \hyperref[eq9]{Eq.~(\ref{eq9})}\;
			
			$\mathbf{g} \leftarrow \text{OGA}(\nabla_T \mathcal{L}) \odot m_{VTG}$\;
			
			$T \leftarrow \text{Update}(T, \mathbf{g})$\;
		}
	}
	\Return $T$\
\end{algorithm}

The training procedure of the proposed method is summarized in \hyperref[Algorithm1]{Algorithm~\ref{Algorithm1}}.

\begin{figure*}[t]
	\centering
	\includegraphics[width=1.0 \textwidth]{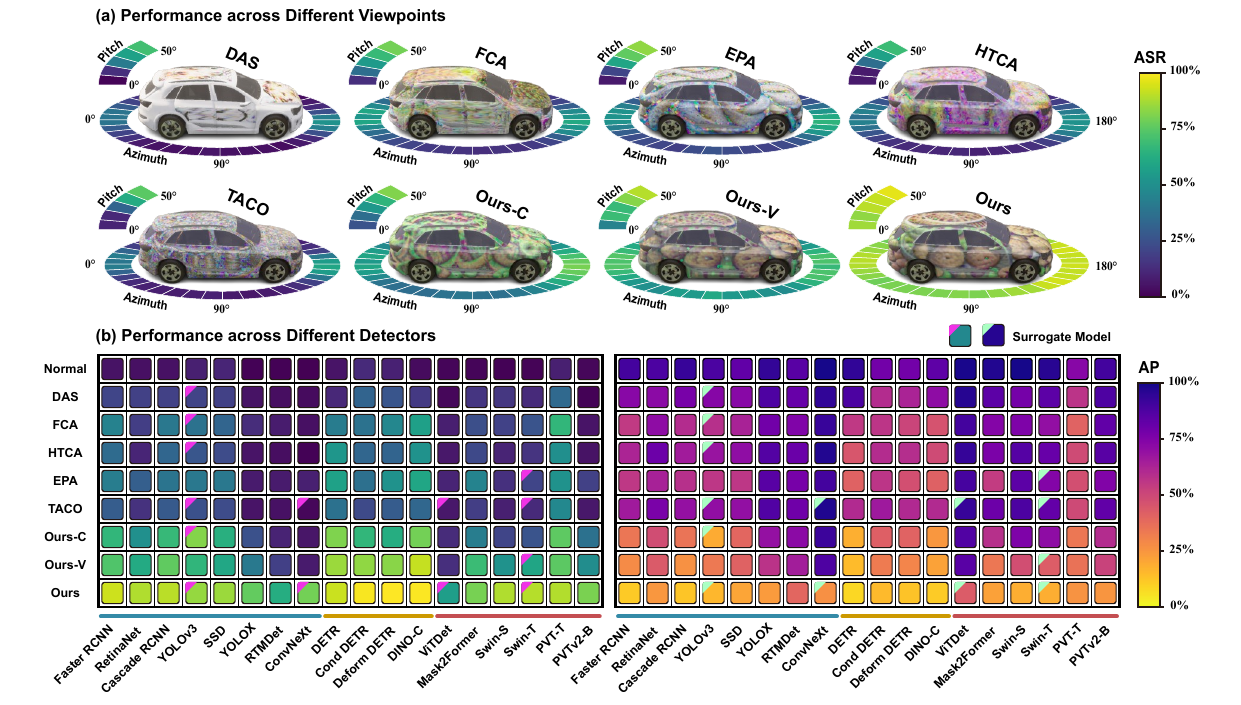}
	\caption{Experimental results in the CARLA simulator, brighter cells indicate stronger attack effects. (a) The central region illustrates the deployment of adversarial textures on 3D entities. The peripheral circular heatmap and the top-left fan-shaped heatmap visualize the average ASR across varying azimuth and pitch angles, respectively, with each grid cell representing a $10^\circ$ span. These ASR values are averaged across all heterogeneous surrogate models to objectively reflect the comprehensive multi-view robustness. (b) Triangular markers in the top-left corners of specific thermal squares denote the corresponding white-box surrogate models utilized by each method.}
	\label{fig8}
\end{figure*}

\section{Experimental Results}
\label{sec4}

\subsection{Implementation Details}
\label{sec41}

\subsubsection{Datasets}
\label{sec411}

To facilitate a fair comparison with prior works and to accurately simulate visual inputs for autonomous driving and security surveillance systems, we utilize the CARLA simulator for dataset generation. We sample 80 distinct locations across multiple maps within the CARLA simulator. Among these, 60 locations are allocated for training physical adversarial attacks, while the remaining 20 are reserved for evaluating attack performance. At each location, we configure four distinct lighting conditions: overcast noon, overcast dawn or dusk, sunny noon, and sunny dawn or dusk. The camera shooting distances are configured at 5-meter intervals, ranging from 5 meters to 40 meters. The camera elevation angles are set from $0^\circ$ to $50^\circ$ in $10^\circ$ increments. Additionally, 36 camera azimuth angles are uniformly sampled at $10^\circ$ intervals starting from $0^\circ$.

\subsubsection{Detectors}
\label{sec412}

To generate adversarial camouflage, we initially train perturbations using white-box models and subsequently evaluate their transferability against black-box models. Following the similarity analysis and greedy selection conducted on the candidate model pool, the white-box detectors selected for JMOF in this study include YOLOv3 \cite{yolov3}, ConvNeXt \cite{ConvNeXt}, Swin-T \cite{Swin_T}, and ViTDet \cite{ViTDet}. The remaining models in the pool are utilized as black-box detectors. Specifically, ConvNeXt, Swin-S, and Swin-T are equipped with the Mask R-CNN detection head \cite{He2017mask}, whereas PVT-T \cite{PVT_T} and PVTv2-B \cite{pvtv2} employ the RetinaNet \cite{RetinaNet} head.

\subsubsection{Training Details}
\label{sec413}

The texture map resolution is set to $512 \times 512$ pixels. The maximum number of training epochs is capped at 10, with a learning rate of 0.01. Furthermore, regarding hyperparameter configurations, the IoU threshold for the detection loss $\mathcal{L}_{det}$ is set to 0.3, and the patch size for the OGA strategy is set to $16 \times 16$ pixels.

\subsubsection{Evaluation Metrics}
\label{sec414}

We evaluate attack performance using Average Precision (AP) and Attack Success Rate (ASR) \cite{Szegedy2013intriguing}. A more substantial reduction in the AP indicates a more thorough disruption of the detector's original capability. Conversely, ASR represents the proportion of samples where the camouflaged target successfully reduces the detector's confidence score below 50\% \cite{Suryanto2022dta,Wang2024generate}.

\begin{figure*}[t]
	\centering
	\includegraphics[width=1.0 \textwidth]{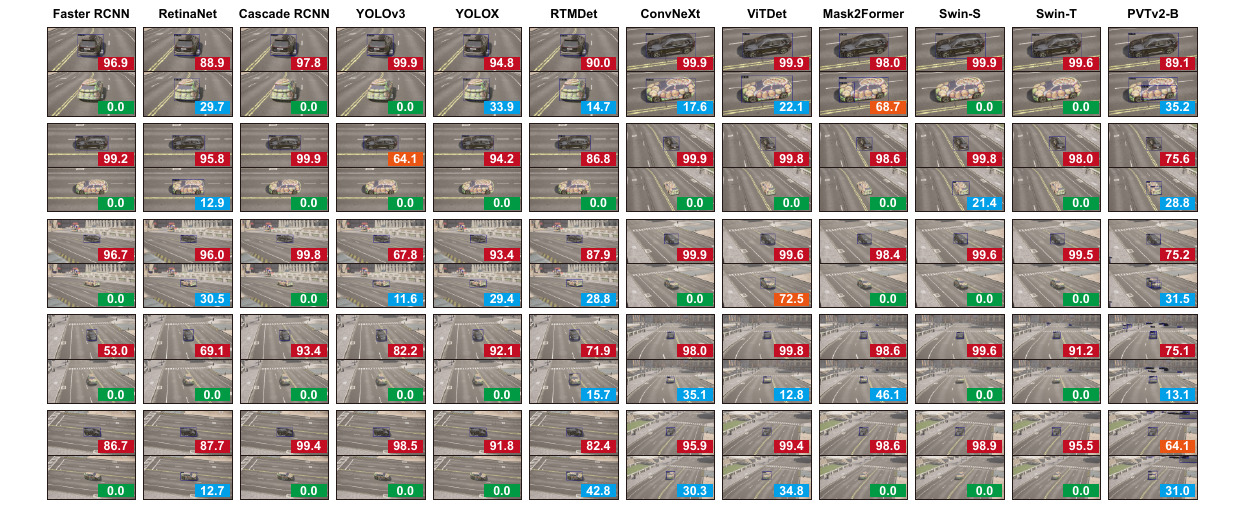}
	\caption{Visualization of representative attack results. In each group, the top row displays the normal target, while the bottom row displays the camouflaged target. The annotations in the bottom-right corners denote the prediction confidence scores output by the detector for the target region. The scores are color-coded as follows: red for 75\%–100\%, orange for 50\%–75\%, blue for 0\%–50\%, and green indicates the complete failure to generate a bounding box for the correct class. This color-coding scheme is consistently applied in all subsequent figures.}
	\label{fig9}
\end{figure*}

\subsection{Universality Evaluation}
\label{sec42}

\subsubsection{Experimental Results in Simulators}
\label{sec421}

As illustrated in \hyperref[fig8]{Fig.~\ref{fig8}}-(a), from a trend analysis perspective, at viewpoints where the azimuth angle approaches $0^\circ$ and $180^\circ$, and the elevation angle approaches $50^\circ$, specifically targeting the front, rear, and roof of the vehicle, the ASR of all camouflage methods remains relatively high. However, when the azimuth angle approaches $90^\circ$, and the elevation angle decreases to $0^\circ$, targeting the vehicle's side, the ASR of most previous methods declines significantly. This degradation is primarily attributed to the sharp expansion of the target's uncamouflaged surface area and abrupt mutations in the 2D perspective topology. In contrast, the proposed method demonstrates the most gradual performance degradation under side-view conditions. This exceptional multi-view robustness is partly attributed to the formidable baseline performance conferred by the JMOF. More importantly, it profoundly validates the necessity of the proposed STD strategy. Traditional optimization tends to take shortcuts, which can easily lead to excessive concentration of adversarial energy in locally compact regions, such as the vehicle's front and rear. This results in a severe absence of adversarial semantics across the large texture areas on the sides. Conversely, the STD strategy forces the algorithm to continuously uncover additional visual vulnerabilities under the dynamic constraints imposed by repeatedly missing local information, and ensuring omnidirectional robustness of the physical adversarial camouflage under complex 3D rotations.

Observing the distribution trends in \hyperref[fig8]{Fig.~\ref{fig8}}-(b), traditional methods optimized on a single surrogate model, such as FCA, DAS, and HTCA, that rely on YOLOv3, alongside EPA, which relies on Swin-T, all exhibit severe limitations due to decision boundary overfitting. Although these methods suffer from a lower overall baseline performance due to the constraints of simple differentiable renderers, they retain a certain deceptive capability against their white-box surrogate models and homologous architecture networks. However, when encountering heterogeneous black-box models with lower similarity, their attack efficacy declines precipitously. More illuminating is that, even though our single-model variants, Ours-C and Ours-V, achieve exceedingly high physical attack baseline performance by leveraging the OPDR rendering pipeline, they also fail to overcome the underlying cognitive barriers of a single model. Their performance remains lackluster when confronting highly heterogeneous networks such as ConvNeXt and RTMDet \cite{RTMDet}. These experimental results demonstrate that relying solely on upgrades to rendering technologies or the deep mining of single-model features fundamentally fails to achieve genuine universal adversarial attacks, thereby compellingly supporting the core motivation and methodology of this paper.

To intuitively evaluate the practical deployment effects of the proposed method, we visualize multiple sets of typical physical adversarial camouflage cases in \hyperref[fig9]{Fig.~\ref{fig9}}. These test cases cover a wide range of camera observation azimuths and diverse physical shooting distances. Under these complex, uncontrollable physical sampling conditions, the adversarial attacks generated by our method consistently and stably deceive detectors. These results visually and compellingly corroborate the multi-view robustness and cross-model transferability of the proposed method within real-world physical scenarios.

\begin{figure}[t]
	\centering
	\includegraphics[width=3.3 in]{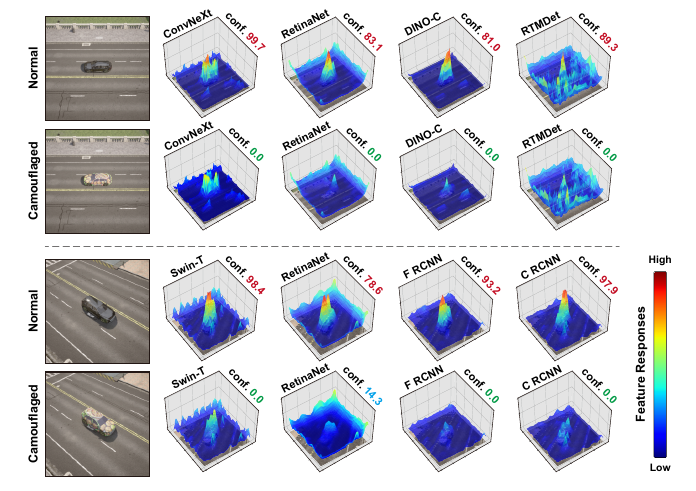}
	\caption{Examples of feature response variations after the detector is attacked. The detector names are displayed in the top-left corners of the feature maps, while the output confidence scores are shown in the top-right corners.}
	\label{fig10}
\end{figure}

Further, we extract and visualize the detector's deep feature response maps when confronting both normal and camouflaged vehicles, as illustrated in \hyperref[fig10]{Fig.~\ref{fig10}}. Observations indicate that for an unperturbed normal vehicle, detectors successfully extract highly structured semantic representations within the target region. However, for vehicles deployed with adversarial camouflage, alongside erroneous detection results, the feature responses exhibit an overall degradation and peak dispersion. This phenomenon clearly demonstrates the efficacy of JMOF's multi-objective design.

\begin{figure}[t]
	\centering
	\includegraphics[width=3.3 in]{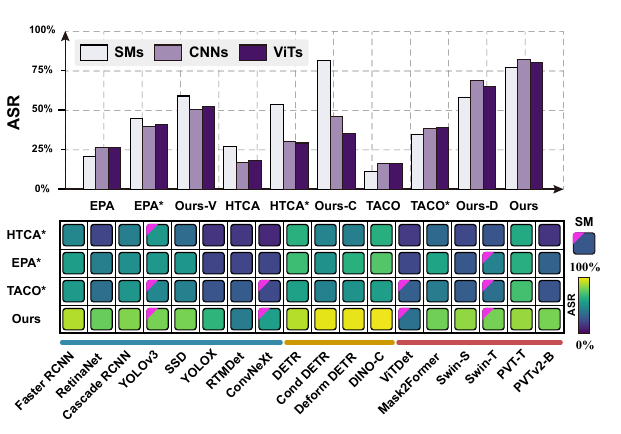}
	\caption{Method comparison after isolating the influence of differentiable renderer performance discrepancies.  The asterisk (*) denotes results obtained by replacing the original differentiable renderer of the method with OPDR.}
	\label{fig11}
\end{figure}

\subsubsection{Performance Comparison of Different Method Designs}
\label{sec422}

As indicated by previous research \cite{NPAC}, during the optimization of physical adversarial attacks, the capacity of the differentiable renderer to reconstruct real physical scenes largely determines the upper bound of attack performance. To isolate the interference of renderer performance discrepancies on attack effectiveness, we uniformly port several classic physical adversarial attack methods onto the OPDR rendering pipeline for re-optimization and evaluation. To ensure a fair comparison in the experiments, alongside Ours-V and Ours-C, we incorporate Ours-D, a variant explicitly excluding the feature loss. The test results, as illustrated in \hyperref[fig11]{Fig.~\ref{fig11}}, demonstrate that although these methods achieve enhanced ASRs leveraging the OPDR pipeline, their performance still falls short of our proposed approach, inherently constrained by their underlying loss functions and optimization mechanisms.

It is particularly noteworthy that, under identical settings, the Swin-T-based Ours-V exhibits superior generalizability against CNNs compared to the YOLOv3-based Ours-C. While potentially counterintuitive in the context of other physical adversarial attack studies, this outcome is highly rational because the Swin-T network has a higher cosine similarity with the other Convolutional Neural Networks in the candidate pool than YOLOv3 does. This clearly demonstrates the necessity of conducting a rigorous similarity analysis before selecting surrogate models.

\subsubsection{Performance Comparison of Different Gradient Fusion Strategies}
\label{sec423}

To evaluate the applicability of existing digital-domain gradient fusion methods in restricted physical parameter spaces, we uniformly port them into our proposed JMOF and OPDR pipeline for fair comparison.

\begin{figure}[t]
	\centering
	\includegraphics[width=3.3 in]{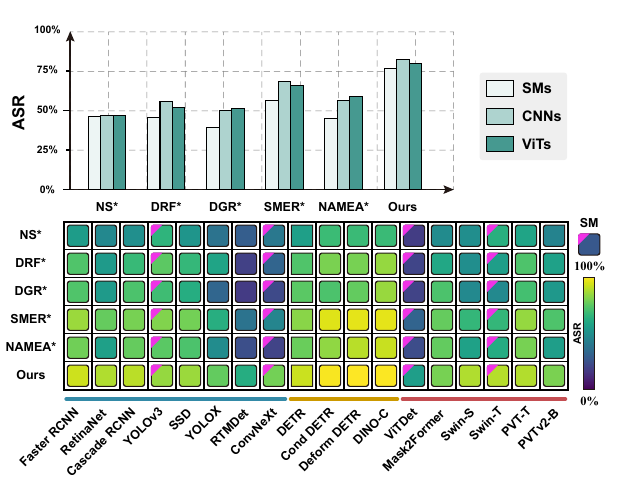}
	\caption{Performance comparison of different gradient fusion strategies. Methods marked with (*) adapt digital gradient strategies for our physical attack workflow.}
	\label{fig12}
\end{figure}

\begin{figure}[t]
	\centering
	\includegraphics[width=3.3 in]{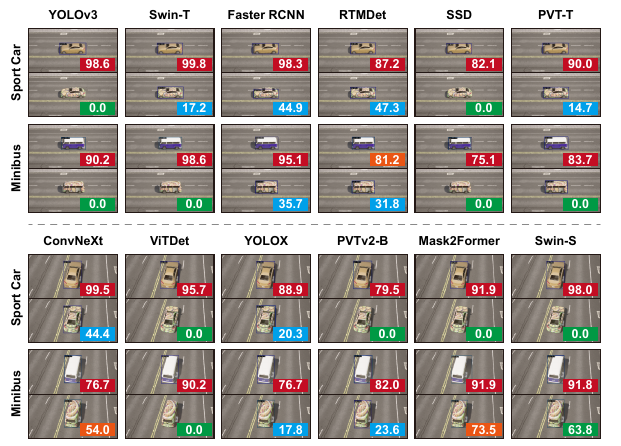}
	\caption{Visualization of representative results after deploying the camouflage onto other heterogeneous targets. In each group, the top row displays the clean target, while the bottom row displays the camouflaged target. Note that only a new UV mapping was applied to the adversarial texture, without any retraining against these heterogeneous models.}
	\label{fig13}
\end{figure}

As illustrated in \hyperref[fig12]{Fig.~\ref{fig12}}, existing methods expose severe performance degradation under complex physical constraints. The DRF and DGR methods essentially adopt a strategy of passive discarding and directional compromise. Upon detecting inconsistent gradient directions, they directly discard the conflicting model gradient components. This retreat mechanism irreversibly and erroneously eliminates the critical feature preferences unique to heterogeneous networks, leading to severe energy dissipation and update stagnation in later stages of joint optimization. Meanwhile, SMER relies on stochastic alternating updates. Within the restricted 3D physical texture space, this strategy triggers exceptionally violent optimization oscillations, failing to achieve stable, common convergence across the decision boundaries of multiple models. Furthermore, although NAMEA attempts to utilize two-dimensional attention masking to separate heterogeneous features during ensemble optimization, it proactively abandons a substantial number of optimizable pixels within the already highly restricted texture map space. This destructive masking intervention severely disrupts the continuity of the underlying physical semantics, leading to a drastic reduction in adversarial energy and plunging adversarial optimization into a fragmentation dilemma.

\begin{figure}[t]
	\centering
	\includegraphics[width=3.0 in]{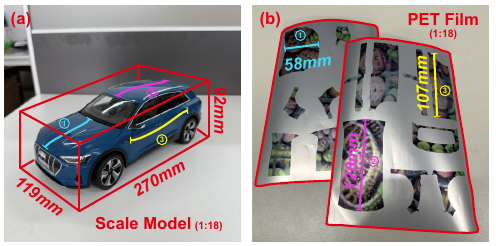}
	\caption{Experimental setup for the physical adversarial attacks. (a) The scaled-down target to be camouflaged. (b) The camouflage carrier material.}
	\label{fig14}
\end{figure}

\begin{figure}[t]
	\centering
	\includegraphics[width=3.3 in]{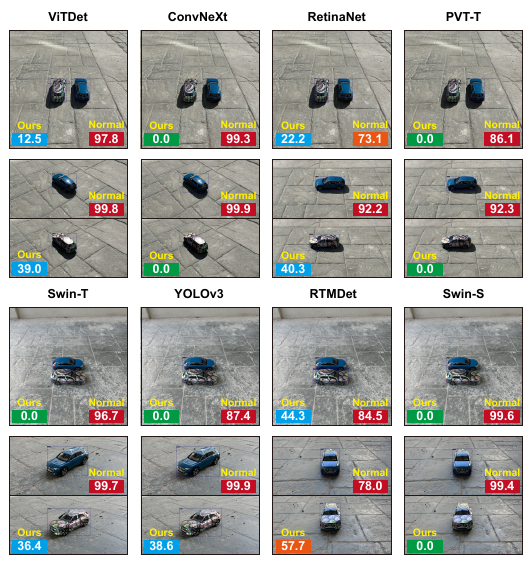}
	\caption{Visualization of representative results from the physical experiments. The annotations in the bottom-left corners denote the confidence scores of the camouflaged target being detected as the correct class, while the bottom-right corners show the confidence scores for the normal target.}
	\label{fig15}
\end{figure}

In contrast, the proposed OGA strategy achieves a paradigm shift in the underlying mathematical logic, transitioning from passive evasion or discarding to active orthogonal restructuring. By accurately capturing and quantitatively resolving local geometric conflicts within the high-dimensional feature space in a patch-wise manner, OGA successfully transforms originally mutually repulsive heterogeneous gradients into synergistic attack kinetic energy, driving the convergence of physical adversarial attacks via spatial projection, achieving this without sacrificing any model diversity features or resorting to energy discarding. This mechanism successfully circumvents the performance degradation suffered by digital attack gradient strategies within the restricted parameter space of physical attacks, thereby breaking the upper bound on ensemble attack performance in heterogeneous generalization.

\subsubsection{Performance on Different Targets}
\label{sec424}

To evaluate the cross-target generalizability of the camouflage, we designed a cross-target attack experiment, directly deploying the camouflage optimized for an SUV model onto the surfaces of heterogeneous targets, such as a sports car and a minibus. As illustrated in \hyperref[fig13]{Fig.~\ref{fig13}}, despite significant structural discrepancies and variations in surface curvature, the camouflage deployed on these heterogeneous targets maintains potent attack effectiveness. This demonstrates that the camouflage generated by our method does not rely on the geometric structure of a specific target; instead, it successfully captures the high-dimensional intra-class semantic features shared across heterogeneous targets, thereby possessing broader applicability and practical deployment value.

\subsubsection{Experimental Results in Physical World}
\label{sec425}

The experiments in this paper that migrate the adversarial attack from the simulator to the real world use the configuration established in our previous work. As illustrated in \hyperref[fig14]{Fig.~\ref{fig14}}, a 1:18-scale authentic factory model is used as the target for camouflage, and aluminized matte PET film is employed as the carrier for the adversarial camouflage. These physical experiments are conducted across diverse angles, varying distances, and different environmental conditions.

In physical-world attack experiments, targets camouflaged by our proposed method achieve ASRs of 65.71\%, 72.76\%, and 70.63\% against SMs, CNNs, and ViTs, respectively. Typical results are presented in \hyperref[fig15]{Fig.~\ref{fig15}}. Compared to performance in the CARLA simulator, the ASR of our method decreases by approximately 10\%. This decline is primarily due to the fact that the 3D scenes in CARLA do not precisely replicate the experimental scenarios of the physical attacks. In future work, it is imperative to further investigate and leverage OPDR to accurately replicate real-world experimental environments.

\begin{table*}[t]
	\centering
	\caption{Ablation study results evaluating the impact of different losses and modules.}
	\label{table1}
	\setlength{\tabcolsep}{5.4 mm}
	\begin{tabular}{cccccc!{\color{gray!90}\vrule width 0.5pt}ccc}
		\toprule
		& \multicolumn{3}{c}{Proposed Losses} & \multicolumn{2}{c}{Proposed Modules}{\color{gray!90}\vrule width 0.5pt}& \multicolumn{3}{c}{ASR} \\
		\cmidrule(lr){2-4} \cmidrule(lr){5-6} \cmidrule(lr){7-9}
		& $\mathcal{L}_{det}$ & $\mathcal{L}_{fea}$ & $\mathcal{L}_{smo}$ & VTG & STD & SMs & CNNs & ViTs \\
		\midrule
		(a) & \checkmark & \checkmark & \checkmark & \checkmark & \checkmark & \textbf{76.81} & \textbf{82.08} & \textbf{79.99} \\
		(b) & & \checkmark & \checkmark & \checkmark & \checkmark & 40.71 & 41.68 & 46.95 \\
		(c) & \checkmark & & \checkmark & \checkmark & \checkmark & 57.91 & 68.75 & 64.93 \\
		(d) & \checkmark & \checkmark & & \checkmark & \checkmark & 63.67 & 67.90 & 66.45 \\
		(e) & \checkmark & \checkmark &\checkmark & & \checkmark & 66.56 & 73.56 & 70.33 \\
		(f) & \checkmark & \checkmark & \checkmark & \checkmark & & 69.97 & 75.85 & 73.42 \\
		\bottomrule
	\end{tabular}
\end{table*}

\subsection{Ablation Studies}
\label{sec43}

\subsubsection{Losses and Modules}
\label{sec431}

We first ablate the proposed loss functions and functional modules to quantify their contributions, with results detailed in \hyperref[table1]{Table~\ref{table1}}. In Experiment (a), the full method achieves ASRs of 76.81\%, 82.08\%, and 79.99\% against Surrogate Models, CNNs, and heterogeneous ViTs, respectively, establishing a robust baseline.

Experiment (b) reveals a precipitous decline in attack performance upon removing the detection loss $\mathcal{L}_{det}$, with ASR across various models plummeting to the 40\% range. This indicates that targeted suppression based on the models' direct outputs constitutes the most fundamental driving force guiding the convergence of adversarial attacks. Without direct penalties at the output end, the optimization process lacks a clear attack direction.

\begin{figure}[t]
	\centering
	\includegraphics[width=3.3 in]{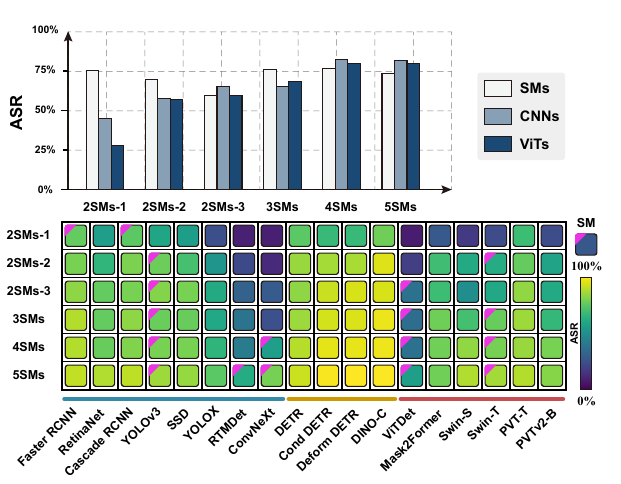}
	\caption{Ablation study results for surrogate model selection within JMOF.}
	\label{fig16}
\end{figure}

Experiment (c) removes the feature loss $\mathcal{L}_{fea}$. Although the model maintains a baseline attack performance, its generalizability across heterogeneous black-box models deteriorates, with the ASR dropping by over 13 percent for CNNs and over 15 percent for ViTs. These data strongly corroborate our previously proposed theory: relying solely on output-layer interventions can easily lead to overfitting in the decoders of surrogate models. By proactively performing mean suppression and variance flattening on feature responses within the deep space via $\mathcal{L}_{feat}$, the method fundamentally dismantles the underlying visual structural representations shared across different networks, thereby assisting the attack in enhancing cross-model transferability.

Experiment (d) removes the smooth loss $\mathcal{L}_{smo}$. The experimental results also demonstrate significant ASR degradation, indicating that the smooth loss is not solely for visual continuity. More importantly, it suppresses false optimization induced by high-frequency adversarial noise and reduces energy dissipation during viewpoint transformations. Simultaneously, it shows that smooth, continuous physical textures can better resist environmental interference, thereby maintaining robustness against physical attacks.

Experiments (e) and (f) ablate the VTG and STD modules, resulting in ASR reductions of 9\% and approximately 6\%, respectively. These performance declines directly substantiate the effectiveness of both enhancement strategies. As detailed in the methodology, VTG prevents the dissipation of adversarial energy into occluded regions to guarantee stable multi-view convergence, whereas STD mitigates pathological reliance on specific local features to ensure robust generalization against unpredictable physical environmental interferences.

\subsubsection{Surrogate Models of JMOF}
\label{sec432}

\hyperref[fig16]{Fig.~\ref{fig16}} presents the ablation on the surrogate model ensemble within JMOF. By varying the composition and quantity of surrogate models, we verify the necessity of gradient similarity pre-analysis and analyze the marginal effects of multi-source joint optimization.

Regarding the selection of surrogate model combinations, a horizontal comparison of different variants comprising two surrogate models reveals that the 2SMs-1 combination, utilizing Faster R-CNN and Cascade R-CNN as surrogate models, exhibits a black-box cross-model attack performance far inferior to the 2SMs-2 and 2SMs-3 versions containing heterogeneous networks such as the YOLOv3, Swin-T, or ViTDet. This discrepancy arises because Faster R-CNN and Cascade R-CNN are highly similar in their underlying architectures and decoding logic, leading to a high degree of overlap between the high-dimensional features they extract and the adversarial gradients generated from them. This experimental result indicates that, within a restricted physical parameter space, blindly combining highly similar homologous models not only fails to expand the generalization boundaries of adversarial camouflage but also causes severe internal resource depletion. Only by introducing heterogeneous surrogate models with low gradient cosine similarity and complementary feature capture preferences can the coverage of adversarial features be fundamentally broadened, thereby enabling robust attacks across network architectures.

In terms of the number of surrogate models, as the number gradually increases from 2 to 4, the cross-model transferability of the adversarial attack shows a distinct stepwise improvement. This sufficiently demonstrates that introducing multidimensional heterogeneous surrogate models can effectively overcome information cognitive barriers and decision biases unique to a single or a few networks, assisting the optimization process in discovering common vulnerable spaces shared by multiple models. However, it is particularly noteworthy that when adding a fifth surrogate model, RTMDet, the overall performance improvement is minimal. From the perspective of gradient similarity analysis, the fundamental reason is that the gradient distribution space of RTMDet is already well covered by existing surrogate models, leaving little incremental adversarial information for the optimizer. This phenomenon of diminishing marginal returns corroborates the core perspective of this paper: the performance upper bound of multi-model joint attacks does not depend on the sheer accumulation of surrogate models, but rather on the orthogonality and complementarity of the model ensemble within the gradient space. Therefore, using the model gradient cosine similarity as a quantitative basis for selecting and eliminating a surrogate model ensemble is indispensable for constructing a highly energy-efficient and highly generalizable joint multi-objective attack framework.

\subsubsection{Patch Size of OGA}
\label{sec433}

Finally, \hyperref[table2]{Table~\ref{table2}} details the ablation on the OGA patch size to analyze the trade-off between local gradient conflict resolution granularity, attack performance, and computational efficiency. Without patching, or with large patches such as $128\times128$ and $64\times64$, overall attack performance remains low despite minimal computational overhead. This is because large-scale orthogonal alignment tends to average local high-frequency gradients with irrelevant low-frequency ones, masking the local semantic divergences that genuinely influence model decisions. As the patch size decreases to $32\times32$ and $16\times16$, the ASR exhibits a steady upward trend while optimization time remains reasonable. However, further refining the patch scale to $8\times8$ displays diminishing marginal returns in attack performance, while optimization time surges exponentially to 5.23 seconds due to the massive increase in computational cost per iteration.

It is worth noting that the optimal patch size is not fixed; it is frequently constrained by the dimensions and proportions of the mesh texture map. An ideal patch setting must precisely match the physical scale at which deep networks extract local features, such as the tire edges and window corners of a target vehicle. An appropriate granularity not only prevents large-scale smoothing but also effectively preserves the integrity of local semantic clusters.

\begin{table}[t]
	\centering
	\caption{Ablation study results on the patch size of OGA.}
	\label{table2}
	\begin{tabular}{ccccc}
		\toprule
		\multirow{2}{*}{\makecell{Patch size}} & \multirow{2}{*}{\makecell{Optimization \\ Time (s)}} & \multicolumn{3}{c}{ASR $\uparrow$} \\
		\cmidrule(lr){3-5}
		& & SMs & CNNs & ViTs \\
		\midrule
		No patch       & 1.27 & 62.90\% & 70.95\% & 73.72\% \\
		128$\times$128 & 1.29 & 66.45\% & 73.22\% & 74.56\% \\
		64$\times$64   & 1.30 & 67.92\% & 75.26\% & 76.41\% \\
		32$\times$32   & 1.42 & 70.81\% & 76.01\% & 76.83\% \\
		16$\times$16   & 1.89 & \textbf{76.81}\% & 82.08\% & \textbf{79.99}\% \\
		8$\times$8     & 5.23 & 76.55\% & \textbf{82.15}\% & 79.85\% \\
		\bottomrule
	\end{tabular}
\end{table}

\subsection{Cross-Task Generalization}
\label{sec44}

To verify the cross-task generalization of JMOF and OGA, we extend our experiments beyond object detection.

\subsubsection{Semantic Segmentation}
\label{sec441}

In cross-task attack experiments, we aim to generate physical adversarial camouflage that simultaneously blinds object detectors and semantic segmentation networks. To this end, we select the object detection model Swin-T and the semantic segmentation model KNet as the surrogate models for joint training. Regarding the design of the loss function, because the semantic segmentation task is fundamentally a pixel-level dense classification, its underlying visual feature extraction logic shares certain similarities with object detection. Therefore, for the semantic segmentation branch, while maintaining the overall framework unchanged, we replace the original detection loss $\mathcal{L}_{det}$ with a specifically designed segmentation loss $\mathcal{L}_{seg}$. Unlike the sparse bounding box-level output of object detection, a semantic segmentation network outputs pixel-wise class probabilities across the entire image. To achieve precise cross-task gradient backpropagation without disrupting the computational graph, we directly extract two-dimensional probability distribution features from the final output of the segmentation network decoder, accurately isolating the dense probability response map for the target category, such as vehicles, denoted as $P_{target}$. Subsequently, as shown in \hyperref[eq13]{Eq.~(\ref{eq13})}, we integrate the 2D mask of the target to be camouflaged, denoted as $m_{target}$, provided by the physical rendering pipeline to apply a globally uniform L1 mean suppression to the valid pixel regions where the physical entity is located. 

\begin{equation}
	\mathcal{L}_{seg}=\frac{\sum(P_{target}\odot m_{target})}{\sum m_{target}+\epsilon}
	\label{eq13}
\end{equation}

\noindent This design drives the target's classification probability toward zero. Global mean suppression prevents vanishing gradients in dense predictions and promotes uniform adversarial coverage across the 3D entity.

\begin{table*}[t]
	\centering
	\caption{Cross-vision-task attack results after joining object detection and semantic segmentation models within JMOF.}
	\label{table3}
	\setlength{\tabcolsep}{4.0 mm}
	\begin{tabular}{cccccccc}
		\toprule
		\multirow{2}{*}{SMs} & \multirow{2}{*}{Method} & \multicolumn{3}{c}{Object Detection (AP$\downarrow$)} & \multicolumn{3}{c}{Semantic Segmentation (IoU$\downarrow$)} \\
		\cmidrule(lr){3-5} \cmidrule(lr){6-8}
		& & Swin-T & Dino-C & Cascade RCNN & KNet & SAN & MaskFormer \\
		\midrule
		\multicolumn{2}{c}{Normal Car} & 93.78 & 80.67 & 89.77 & 83.80 & 84.41 & 85.14 \\
		Swin-T \& YOLOv3 & OGA & 41.49 & 16.56 & 21.84 & 71.39 & 80.44 & 54.66 \\
		Swin-T \& KNet   & NS  & 57.12 & 37.42 & 41.22 & 59.35 & 77.63 & 36.57 \\
		Swin-T \& KNet   & OGA & \textbf{46.59} & \textbf{24.16} & \textbf{28.84} & \textbf{37.91} & \textbf{63.20} & \textbf{27.05} \\
		\bottomrule
	\end{tabular}
\end{table*}

\begin{figure}[t]
	\centering
	\includegraphics[width=3.3 in]{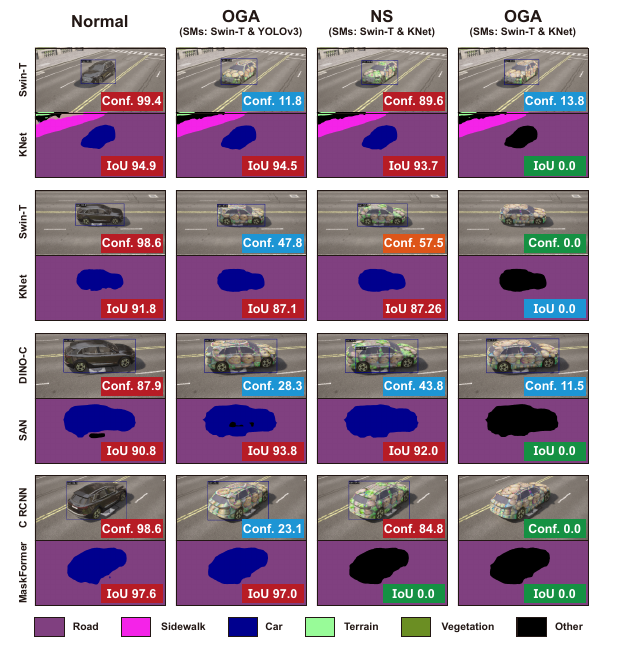}
	\caption{Visualization of representative attack results across object detection and semantic segmentation tasks.}
	\label{fig17}
\end{figure}

As shown in \hyperref[table3]{Table~\ref{table3}} and \hyperref[fig17]{Fig.~\ref{fig17}}, when forcibly integrating these distinct tasks within a restricted physical texture space, traditional equal-weight summation strategies trigger severe cross-task gradient conflicts, substantially degrading performance on both tasks. However, empowered by OGA's dynamic restructuring, our method successfully resolves this geometric repulsion. The generated camouflage maintains high ASR on object detection while causing a precipitous drop in the target's IoU for segmentation, demonstrating robust cross-task generalization.

\subsubsection{Monocular Depth Estimation}
\label{sec442}

Beyond crossing visual tasks with similar underlying logic, JMOF even exhibits the potential to bridge high-level semantic perception and low-level geometric regression models, which differ drastically in task nature. Regarding the surrogate models, we select the object detection model Swin-T and the monocular depth estimation model DPT. In terms of loss function design, the monocular depth estimation task involves continuous spatial regression, and models with different architectures may output absolute depth maps or relative disparity maps. To ensure that the attack eradicates the target from the model's spatial perception, we construct a depth loss $\mathcal{L}_{dep}$. As formulated in \hyperref[eq14]{Eq.~(\ref{eq14})}, this loss integrates the 2D mask $m_{target}$ of the target to be camouflaged, provided by the physical rendering pipeline, extracts a dense response on the target surface, and applies global L1 suppression using a unified inverse-depth transformation.

\begin{equation}
	\mathcal{L}_{depth}=\frac{\sum((D_{pred}+\epsilon)^{-1}\odot m_{target})}{\sum m_{target}+\epsilon}
	\label{eq14}
\end{equation}

\noindent By minimizing this loss, the algorithm forcibly suppresses the inverse depth values of the target's visible region toward zero, thereby compelling the camouflaged vehicle to blend into the background at the geometric level of perception. Furthermore, it should be noted that during the optimization process targeting the monocular depth estimation model, we do not introduce a feature loss into the depth algorithm. This is because monocular depth networks primarily infer depth from low-frequency global topology and structural priors, exhibiting robust geometric error correction and recovery against local high-frequency feature perturbations. Dispersing the feature layer yields diminishing returns in dismantling its global spatial distance perception; instead, it dilutes the optimization energy and delays convergence.

\begin{figure}[t]
	\centering
	\includegraphics[width=3.3 in]{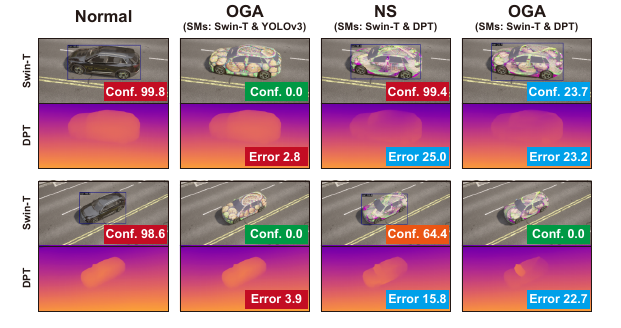}
	\caption{Visualization of representative attack results across object detection and monocular depth estimation tasks.}
	\label{fig18}
\end{figure}

The specific cross-task experimental evaluation results are illustrated in \hyperref[table4]{Table~\ref{table4}} and \hyperref[fig18]{Fig.~\ref{fig18}}. It can be clearly observed that when confronting a task combination with diametrically opposed underlying cognitive logics, the OGA mechanism effectively maintains multi-task attack performance. This outcome strongly validates the exceptional universality and synergistic generalization potential of the proposed multi-source gradient joint optimization framework across heterogeneous visual tasks.

\begin{table}[t]
	\centering
	\caption{Cross-vision-task attack results after joining object detection and depth estimation models within JMOF. Depth Error denotes the discrepancy in depth estimation results of the camouflaged target relative to the normal target.}
	\label{table4}
	\begin{tabular}{cccc}
		\toprule
		SMs & Method & AP $\downarrow$ & Depth Error $\uparrow$ \\
		\midrule
		\multicolumn{2}{c}{Normal Car} & 93.78 & - \\
		Swin-T \& YOLOv3 & OGA & 41.49 & 7.92 \\
		Swin-T \& DPT    & NS  & 64.54 & 20.42 \\
		Swin-T \& DPT    & OGA & \textbf{46.13} & \textbf{56.68} \\
		\bottomrule
	\end{tabular}
\end{table}

\subsection{Discussion}
\label{sec46}

This paper proposes a universal underlying logic capable of quantifying and resolving gradient conflicts across architectures and tasks. This approach endows the method with exceptional scalability. When confronting novel object detectors or visual networks with distinct mechanisms that may emerge in the future, our framework eliminates the need for blind trial and error. Instead, one can precalculate the gradient cosine similarity between the new model and the existing model pool. By quantitatively analyzing the spatial gradient distribution of the new algorithm, we can accurately determine its feature overlap and the intensity of geometric conflicts with existing models at the decision boundaries. This establishes the framework as a powerful diagnostic and fusion tool, adaptively incorporating novel neural network architectures into the joint optimization camp and continuously elevating the generalization upper bound for physical adversarial attacks.

In future research, the core philosophy of this paper can be generalized to a broader space of visual tasks. This study has preliminarily verified the synergistic attack capabilities of the method across object detection, semantic segmentation, and monocular depth estimation tasks. Moving forward, we plan to broaden the application boundaries of JMOF and OGA to more complex real-world visual tasks with stricter security requirements, such as time-series-based trajectory prediction, 3D point cloud detection in autonomous driving, and even facial recognition systems. We aim to explore how to effectively extract and reconstruct more fundamental, universal semantic vulnerabilities within these multimodal or high-security-level tasks.

Although the proposed method achieves breakthrough performance gains when combining heterogeneous models and tasks, this leap inevitably entails increased computational cost. In each iteration of JMOF, the simultaneous execution of forward propagation and backpropagation across multiple complex models results in noticeable increases in both memory consumption and computational time. In subsequent research, we will devote efforts to exploring more efficient joint-optimization algorithms and lightweight multi-model synergistic mechanisms. The objective is to significantly reduce the computational complexity and hardware resource overhead of the algorithm while maintaining or even surpassing the current cross-domain attack efficacy, ultimately achieving a sustainable balance between the generation efficiency and generalization performance of physical adversarial attacks.

\section{Conclusion}
\label{sec5}

This paper delves into the key limitations that constrain the generalization capability of physical adversarial attacks. First, existing methods frequently overfit to single surrogate models, causing the generated camouflage to be confined to specific decision boundaries. Second, current attacks typically lack an effective synergy between shallow prediction suppression and deep feature disruption, hindering their ability to simultaneously achieve high attack efficiency and broad generalizability. Finally, when integrating multiple optimization objectives and models within restricted physical texture spaces, conventional fusion strategies often trigger gradient conflicts and energy depletion, resulting in significant attack degradation.

To address these challenges, JMOF is proposed to generate physical adversarial attacks with enhanced generalization capabilities. By introducing a quantitative analysis of gradient similarity, JMOF defines an effective similarity regime to optimize surrogate model selection, identifying an ensemble that covers a broader adversarial feature space with complementary features. Furthermore, a dual-level multi-objective attack mechanism is designed to jointly suppress prediction outputs and flatten intermediate feature distributions, balancing attack efficiency with deep generalization. Within JMOF, the OGA strategy is introduced to resolve multi-source gradient conflicts, transforming mutually repulsive heterogeneous gradients into synergistic optimization directions that converge on shared vulnerable regions across multiple models. Extensive digital-to-physical experiments demonstrate that the proposed method not only achieves consistent cross-model transferability against diverse object detectors but also exhibits effective cross-vision-task generalization. It generates attacks capable of simultaneously disrupting object detection and semantic segmentation or monocular depth estimation models. This research advances the performance boundaries of physical adversarial attacks, providing a robust framework for evaluating the vulnerability of visual AI systems in real-world deployments.

\bibliographystyle{IEEEtran}
\bibliography{Ref}

\end{document}